\documentclass[sigconf]{acmart}
\AtBeginDocument{%
  }
\usepackage{subfig}
\usepackage{makecell}
\usepackage{multirow} 

\copyrightyear{2025}
\acmYear{2025}
\setcopyright{acmlicensed}\acmConference[SIGIR '25]{Proceedings of the 48th International ACM SIGIR Conference on Research and Development in Information Retrieval}{July 13--18, 2025}{Padua, Italy}
\acmBooktitle{Proceedings of the 48th International ACM SIGIR Conference on Research and Development in Information Retrieval (SIGIR '25), July 13--18, 2025, Padua, Italy}
\acmDOI{10.1145/3726302.3731946}
\acmISBN{979-8-4007-1592-1/2025/07}

\begin{document}

\title{PaRT: Enhancing Proactive Social Chatbots with  Personalized Real-Time Retrieval}

\author{Zihan Niu$^{\heartsuit}$, Zheyong Xie$^{\spadesuit}$, Shaosheng Cao$^{\spadesuit}$$^{\dagger}$, Chonggang Lu$^{\spadesuit}$, Zheyu Ye$^{\spadesuit}$ \\ Tong Xu$^{\heartsuit}$$^{\dagger}$, Zuozhu Liu$^{\diamondsuit}$$^{\dagger}$, Yan Gao$^{\spadesuit}$, Jia Chen$^{\spadesuit}$, Zhe Xu$^{\spadesuit}$, Yi Wu$^{\spadesuit}$, Yao Hu$^{\spadesuit}$}\thanks{$^{\dagger}$Corresponding authors.}
\affiliation{
  \institution{$^{\spadesuit}$Xiaohongshu Inc.\\$^{\heartsuit}$University of Science and Technology of China\\$^{\diamondsuit}$Zhejiang University}
  \country{}
}
\email{niuzihan@mail.ustc.edu.cn, tongxu@ustc.edu.cn, zuozhuliu@intl.zju.edu.cn}
\email{{xiezheyong, caoshaosheng, luchonggang, zheyuye}@xiaohongshu.com}
\email{{yadun, chenjia2, qiete, xiaohui, houxia}@xiaohongshu.com}

\renewcommand{\shortauthors}{Zihan Niu et al.}
\newcommand{\edit}[1]{\textcolor{red}{#1}}
\newcommand{\xzy}[1]{\textcolor{bleu}{#1}}

\begin{abstract}
Social chatbots have become essential intelligent companions in daily scenarios ranging from emotional support to personal interaction. However, conventional chatbots with passive response mechanisms usually rely on users to initiate or sustain dialogues by bringing up new topics, resulting in diminished engagement and shortened dialogue duration. 
In this paper, we present PaRT, a novel framework enabling context-aware proactive dialogues for social chatbots through personalized real-time retrieval and generation.  
Specifically, PaRT first integrates user profiles and dialogue context into a large language model (LLM), which is initially prompted to refine user queries and recognize their underlying intents for the upcoming conversation. Guided by refined intents, the LLM generates personalized dialogue topics, which then serve as targeted queries to retrieve relevant passages from RedNote. Finally, we prompt LLMs with summarized passages to generate  knowledge-grounded and engagement-optimized responses. Our approach has been running stably in a real-world production environment for more than 30 days, achieving a 21.77\% improvement in the average duration of dialogues.
\end{abstract}



\maketitle

\section{Introduction}
Recent advancements in Large Language Models (LLMs) \cite{achiam2023gpt, bai2023qwen, touvron2023llama} have significantly boosted the development of social chatbots \cite{zhou2020design, adiwardana2020towards}, enabling them to exhibit increasingly intelligent and human-like behavior, gathering widespread attention. 
Despite this progress, current social chatbots primarily emphasize providing comprehensive and emotional responses to user requests \cite{10.1145/3539618.3591946, wang2024memory}, neglecting the importance of actively engaging with users in conversation \cite{ijcai2023p738, 10.1145/3626772.3657843, 10.1145/3539618.3594250}.
For instance, traditional chatbots relying solely on passive dialogue strategies \cite{ijcai2023p738, 10.1145/3539618.3594250} often limit the depth and natural extension of conversation, requiring users to continually initiate and drive the conversation, resulting in decreased user engagement and shortened interaction duration.

\begin{figure}
    \centering
    \includegraphics[width=1.0\linewidth]{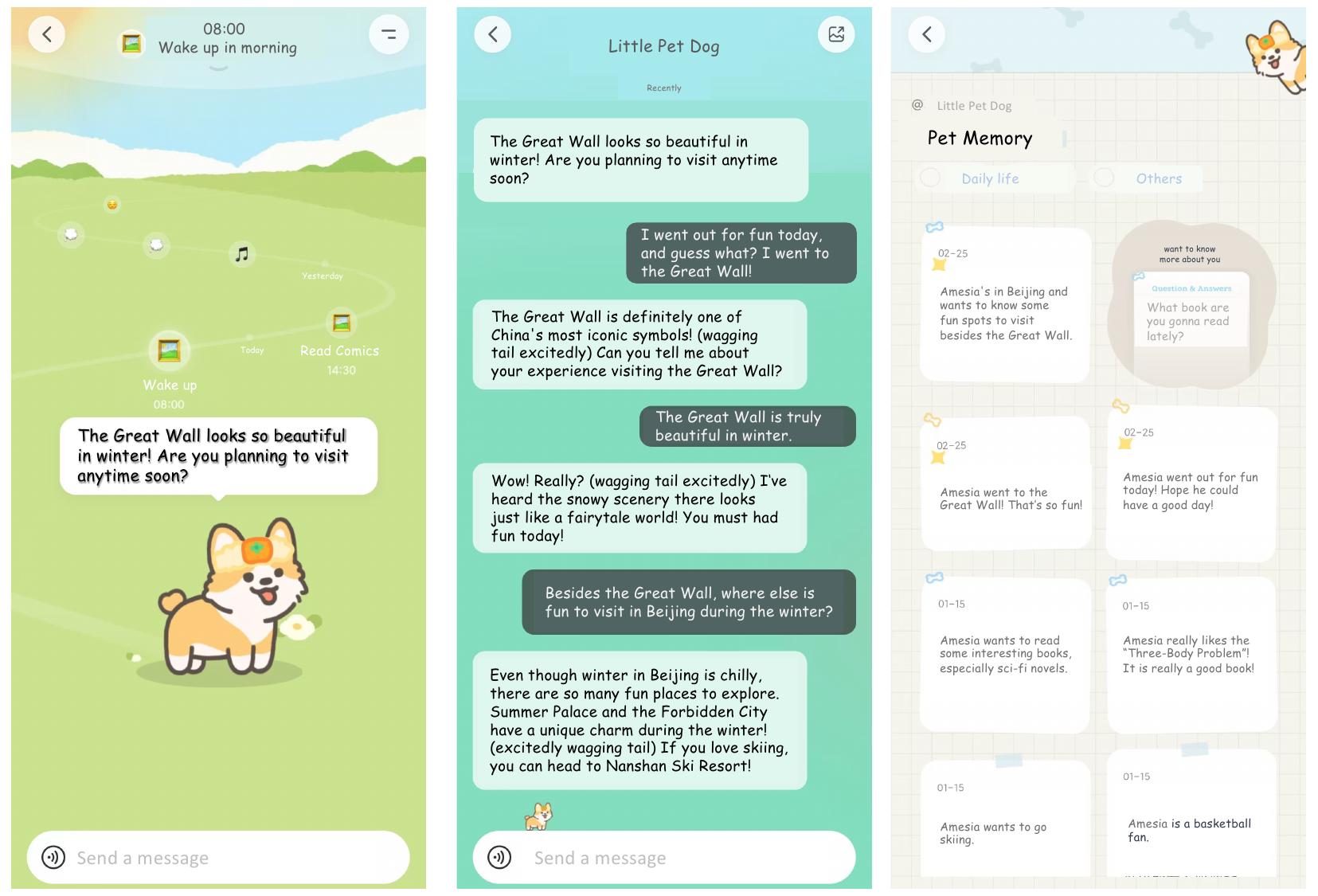}
    \caption{The interface of PaRT. From left to right, we sequentially present the greeting interface, the dialogue page, and the user profile interface. The greeting scenario is initiated by chatbot at the beginning, while the dialogue scenario aims to proactively guide ongoing conversation.}
    \label{fig:interface}
\end{figure}

\begin{figure*}
    \centering
    \includegraphics[width=1.0\linewidth]{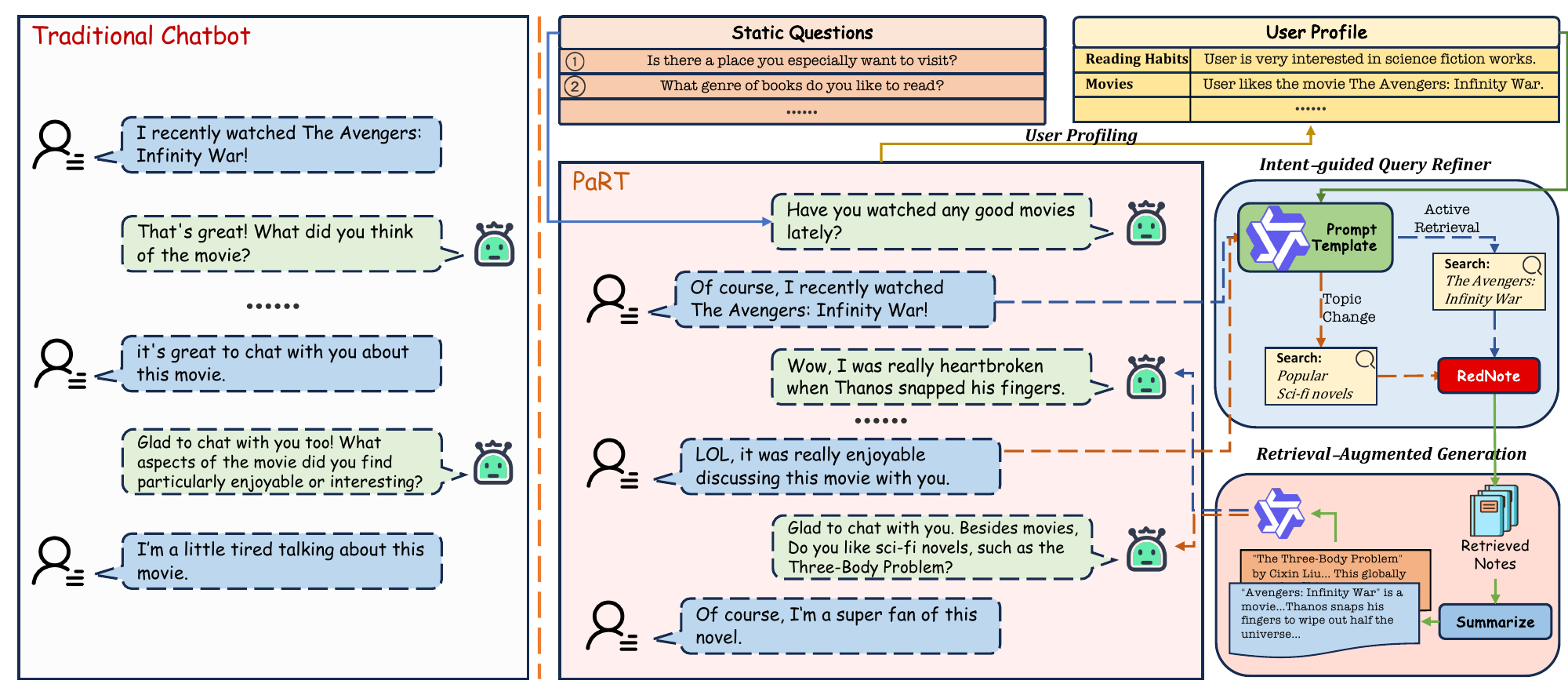}
    \caption{The figure provides an overview of PaRT. It illustrates the different dialogue experiences based on traditional chatbot (left) and our PaRT method (right). }
    \label{fig:overview}
\end{figure*}

Proactive dialogues have emerged as a promising solution to enhance user engagement in conversational AI. Based on established definitions of proactivity~\cite{grant2008dynamics, ijcai2023p738}, these systems actively initiate or switch topics during interactions rather than maintaining passive responsiveness to users. For instance, they may initiate conversations with domain-specific prompts (e.g., "Which tourist attractions interest you most?") or dynamically shift topics when detecting waning user interest~\cite{10.1145/3539618.3591880, 10.1145/3477495.3531844}. 
Current implementations of proactive dialogues often directly prompt LLMs for topic generation, resulting in two critical limitations: 1) Overgeneralized content. Direct prompting usually produces generic topics misaligned with user preferences, degrading interaction quality; 2) Knowledge boundaries: LLMs' inherent limitation in knowledge constrains their ability to sustain domain-specific and up-to-date dialogues with real-time contextual information~\cite{10.1145/3397271.3401296, 10.1145/3626772.3661375}.

To address these challenges, we propose \textbf{PaRT}, which enhances \textbf{P}ro\textbf{a}ctive social chatbots with personalized
real-time \textbf{R}e\textbf{T}rieval.
Our proposed PaRT adopts a unified framework that addresses two essential proactive conversation scenarios, as illustrated in Figure \ref{fig:interface}. Both scenarios are supported by three key components: user profiling, an intent-guided query refiner, and retrieval-augmented generation.
In particular, the user profiling module lays the foundation by constructing a detailed user profile through proactive questioning and memory extraction \cite{zhong2024memorybank, lu2023memochat,10.1145/3715097}.
Subsequently, the intent-guided query refiner analyzes the dialogue context to identify user intent, which includes natural transitions, explicit retrieval (direct information requests), and implicit retrieval (signals for topic change). 
Based on the identified intent and user profile \cite{10.1145/3626772.3657947, 10.1145/3477495.3531789}, it then refines the current query into a personalized one.
This refined query serves as a concrete representation of the topic for the upcoming conversation.
Leveraging the tailored queries, the retrieval-augmented generation module performs retrieval from RedNote\footnote{\url{https://www.xiaohongshu.com}}, and summarizes these retrieved passages to mitigate interference from irrelevant information \cite{lyu2024retrieve, xu2023recomp}.
Then we prompt LLMs with summarized text to generate responses that are well-aligned with the personalized query \cite{lewis2020retrieval, jiang2023active}.
Both offline and online experimental results demonstrate that PaRT significantly enhances generation quality and improves user engagement.

In summary, our contributions are as follows: 
\begin{enumerate}
    \item 
    To the best of our knowledge, this is the first work proposing the enhancement of proactive social chatbots with personalized real-time retrieval, which significantly improves generation quality and effectively enhances user engagement by initiating and guiding conversation topics.
    \item Our method achieves an effective proactive dialogue experience with personalized real-time retrieval by leveraging user profiling, intent-guided query refiner and retrieval-augmented generation.
    \item The proposed PaRT has been successfully deployed in a real-world production environment for over 30 days, achieving remarkable results by increasing the average dialogue duration by 21.77\%.
\end{enumerate}

\section{Details of the Proposed PaRT}
The architecture of PaRT is illustrated in Figure\ref{fig:overview}. 
In this section, we present a detailed explanation of PaRT, including its overall workflow and three core modules: user profiling, intent-guided query refiner, and retrieval-augmented generation. 

\subsection{User Profiling}
User profiling module is first introduced to enhance chatbots' understanding of user preferences
By introducing memory mechanisms \cite{zhong2024memorybank, lu2023memochat}, critical information from dialogue history is summarized and stored in user profiles. 
Additionally, proactive greetings serve as an effective method for profile modeling.
We have developed a comprehensive static question bank from which the chatbot can randomly select questions to initiate conversations. 
Furthermore, personalized greetings based on existing user profiles and retrieval-augmented generation (detailed in Section 2.4) facilitate the development of more comprehensive user profiles.
As illustrated in Figure \ref{fig:overview}, conversations begin with a greeting, while user preferences are continuously updated to maintain real-time user profiles.

\subsection{Intent-guided Query Refiner}
As conversations evolve, effective chatbots should proactively address users' needs or pivot topics when engagement wanes. 
Figure \ref{fig:overview} (left) illustrates how traditional chatbots persist with the current topic despite declining user engagement, resulting in suboptimal interaction quality.
Only by truly understanding users' underlying intents can chatbots drive higher quality conversations.

To identify users' true intents \cite{10.1145/3688400, 10.1145/3700890, 10.1145/3477495.3532069}, we introduce an intent-guided query refiner module based on LLM. 
Specifically, we categorize user intent into three types: natural transition, explicit retrieval, and implicit retrieval.
Natural transition indicates that the chatbot should sustain dialogue while providing companionship.
For explicit retrieval situations (e.g., "What do you think about the recently released Dune 2?"), chatbots must perform active retrieval \cite{jiang2023active, lewis2020retrieval} to access latest information. However, not all user queries explicitly indicate retrieval needs. As conversations evolve, user interactions may signal declining interest or tendencies toward topic shifts \cite{xie2021tiage, rachna2021topic}, implying an implicit retrieval intent. In such cases, chatbots must detect waning user interest and generate personalized topic transitions based on the conversation context and user profiles to maintain engagement. To enhance retrieval effectiveness in both situations, we introduce query rewriting \cite{10.1145/3626772.3657933, peng2024large, 10.1145/3397271.3401323}, which clarifies the target entity and minimizes interference from irrelevant content by leveraging the advanced understanding capabilities of LLMs \cite{achiam2023gpt, bai2023qwen}. We carefully design prompts to guide the LLM to fully consider the conversation context and user profiles, first identifying user intent, then refine current query to match user preferences for subsequent retrieval. These refined queries enable topic-level control while enhancing response quality through information retrieval \cite{lyu2024retrieve}.
The entire intent-guided query refining process can be formally represented as:
\begin{align}
    c, q = LLM(profile, context, prompt_r),
\end{align}
where $c$ denotes intent category, $q$ stands for rewritten query, and $prompt_r$ is the prompt template for intent-guided query refining.

\subsection{Retrieval-Augmented Generation}
Proactive chat scenarios, particularly those involving open-domain lifestyle conversations, greatly benefit from integrating real-time information. In these contexts, static knowledge bases are inherently limited and quickly become outdated \cite{qin2023webcpm, 10.1145/3626772.3657660}, making web search essential for retrieval-augmented generation (RAG). To fully leverage this advantage, we adopt a conventional RAG framework \cite{lewis2020retrieval, jiang2023active, lyu2024retrieve} that operates in three stages: retrieval, summarization, and generation with web sources.
In our practical implementation, we conduct retrieval using the RedNote search engine, a popular lifestyle-sharing platform containing abundant high-quality content that provides real-time insights across diverse topics, thus ensuring retrieval quality.
To ensure consistent generation quality, PaRT employs distinct prompts tailored to different proactive conversation scenarios. For the greeting scenario, we randomly select an entry from the user profile and prompt the LLM to summarize the core interest as the search query. In contrast, for the dialogue scenario, we utilize the query generated by our intent-driven query refiner. 
In both the greeting and dialogue scenarios, the generated query triggers the retrieval operation. An LLM then summarizes the top k retrieved passages based on the query, effectively filtering out irrelevant information \cite{xu2023recomp, 10.1145/3626772.3657834}. This summarized content is subsequently integrated with contextual information, enabling the generation of responses that are both natural and richly informative.
The whole RAG process can be formally represented as:
\begin{align}
    note_1, \dots, note_k &= Retriever(q), \\
    summary &= LLM(q, [note_1, \dots, note_k], prompt_s), \\
    response &= LLM(context, summary, prompt_g),
\end{align}
where $Retriever$ denotes search engine, $note$ stands for retrieved passage, while $prompt_s$ and $prompt_g$ are prompt templates for summarization and generation, respectively.

\section{Experiments}
In this section, we first present the implementation details of PaRT and the evaluation metrics.
Then we conduct comprehensive offline experiments based on a testset derived from iPET\footnote{\url{https://xhslink.com/L8wKw6}}, a social chatbot deployed on RedNote. Finally, we report the online experimental results following the deployment of PaRT in a real-world production environment.

\subsection{Implementation Details}
For model training, we construct a high-quality dataset comprising 11,455 samples and employ Supervised Fine-Tuning (SFT) with full parameter optimization on the Qwen2 family \cite{yang2024qwen2technicalreport}. To balance the latency-cost tradeoff, Qwen2-72B-Instruct is used for dialogue generation, while Qwen2-7B-Instruct is utilized for other components. The model is optimized with a context length of 2048, a learning rate of 5e-6, and a cosine decay schedule, incorporating a 0.1 warmup ratio. The batch size is set to 2 per device, and gradient accumulation is performed with 4 steps to ensure stable training. 
The entire training process consists of 3 epochs, utilizing 24 NVIDIA A100 80GB GPUs, with a total training time of approximately 4 hours.
Additionally, for model inference, a temperature coefficient of 0.9 is applied to balance the creativity and determinism of the output.

\subsection{Evaluation Metrics}
To thoroughly validate the capabilities of PaRT, we evaluate PaRT from both retrieval and generation perspectives.
Given the difficulty of traditional machine methods in evaluating subjective tasks, we use an LLM-based evaluation \cite{zhou2024characterglm, 10.1145/3626772.3661346}, which has shown alignment with human judgment \cite{chiang2023closer, chiang2023can}.
For each task, we randomly sample 50 examples and report the kappa agreement \cite{sim2005kappa, 10.1145/3397271.3401334} between LLM and human ratings.
To evaluate retrieval performance, we only adopt the \textbf{Precision@k} (P@k) metric \cite{zhu2004recall} since the retrieval database is large and dynamic. 
We design prompts and ask LLM to verify whether retrieved passages meet all three requirements: relevance, usefulness, and conversational coherence, assigning labels of \texttt{0} or \texttt{1}.
Furthermore, we evaluate generation quality in both greeting and dialogue scenarios using subjective metrics inspired by \cite{cheng2023pal, tu2024charactereval}, focusing on three dimensions: \textbf{Personalization}, \textbf{Informativeness}, and \textbf{Communication Skills}.
The personalization metric assesses how well the response adapts to the user's preferences to reflect individualized relevance, while the informativeness metric measures the richness of information contained in the response. And the communication skills metric evaluates the coherence, emotional resonance, and engagement of the response in fostering natural and meaningful user interactions.
To quantify these aspects, we require the LLM to assign a score for each dimension, ranging from \texttt{0} to \texttt{3}, with higher scores denoting better quality.

\subsection{Offline Experimental Results}

\subsubsection{Retrieval Performance}
We compare the retrieval performance between methods based on original user query and LLM-rewritten query.
As shown in Table \ref{tab:retrieval}, using the rewritten query method leads to a 31.71\% improvement in overall retrieval performance compared to the user query method.
Additionally, the table indicates that P@k of rewritten queries declines slower as more passages are retrieved, highlighting the robustness of our approach.
These experimental results support conclusions similar to \cite{10.1145/3626772.3657933, ye2023enhancing}, indicating that LLM-based query rewriting significantly enhances retrieval performance.
To optimize the balance between information richness and retrieval precision, we set the number of retrieved passages to 5.

\begin{table}[htbp]
  \caption{Retrieval performance on offline dataset}
  \label{tab:retrieval}
  \centering
  \renewcommand{\tabcolsep}{3pt}
  \begin{tabular}{lccccc}
    \toprule
    \textbf{Method} & \textbf{P@1} & \textbf{P@3} & \textbf{P@5} & \textbf{P@10} & \textbf{Avg} \\
    \midrule
    User Query & 0.5871   & 0.4428   & 0.3400 & 0.2001 & 0.3925 \\
    Rewritten Query & \textbf{0.7847}   &  \textbf{0.7361}   & \textbf{0.7056} & \textbf{0.6121} &  \textbf{0.7096}\\
    \midrule
    \multicolumn{5}{l}{Kappa (vs. Human): 0.78}\\
    \bottomrule
  \end{tabular}
\end{table}

\subsubsection{Generation Performance}

To evaluate the generation quality of our approach, we conduct a comprehensive analysis using three distinct metrics, as illustrated in Section 3.2. 
In our experiments, we compare two methods: direct generation and personalized generation. 
Direct generation is the traditional chatbot approach using LLM responses immediately, while personalized generation approach produces responses based on user profiles without external information.
PaRT goes further by utilizing both user profiles and retrieved notes.
Table \ref{tab:response} summarizes the detailed experimental results. 
The results demonstrate that personalized generation provides more satisfying responses compared to direct generation method, confirming the significant value of user profiles in enhancing generation quality, as also validated in \cite{cheng2023pal}. 
Notably, PaRT outperforms other methods across all evaluation metrics, substantiating that incorporating personalized retrieval into the generation process substantially improves response quality, delivering more tailored interactive experiences to users.

\begin{table}[htbp]
    \caption{Generation performance on offline dataset}
    \label{tab:response}
    \centering
    \begin{tabular}{lcccc}
        \toprule
        \textbf{Method} & \textbf{Pers.} & \textbf{Info.} & \textbf{Coms.} & \textbf{Avg} \\
        \hline
        \textit{Greeting} \\
        \hline
        Direct Generation & 0.9094 & 1.1353 & 1.5381 & 1.1943 \\
        Persona Generation & 1.2290 & 1.9334 & 2.1624 & 1.7749 \\
        PaRT & \textbf{1.7534} & \textbf{2.0978} & \textbf{2.4090} & \textbf{2.0867}  \\
        \hline
        \textit{Dialogue}\\
        \hline
        Direct Generation & 1.4554 & 1.7183 & 2.2075 & 1.7937 \\
        Persona Generation & 1.5684 & 1.8722 & 2.1940 & 1.8782 \\
        PaRT & \textbf{2.1592} & \textbf{2.1484} & \textbf{2.3098} & \textbf{2.2058} \\
        \hline
        Kappa (vs. Human) & 0.51 & 0.46 & 0.43 & - \\
        \bottomrule
    \end{tabular}
\end{table}

\subsubsection{Impact of Retrieval Quantity on PaRT Performance}
We further investigate the impact of retrieval quantity k on PaRT by evaluating four configurations (k=1, 3, 5, and 10) across greeting and dialogue scenarios. As shown in Table \ref{tab:size k}, k=5 yields optimal performance in both scenarios, while less notes provide insufficient information and excessive notes introduce performance-degrading noise.
These results align with the previous observation from \cite{wang2024astute} that effective RAG systems must carefully balance between retrieval comprehensiveness and precision.

\begin{table}[htbp]
    \caption{Impact of retrieval quantity in PaRT}
    \label{tab:size k}
    \centering
    \begin{tabular}{lcccc}
        \toprule
        \multirow{2}{*}{\textbf{Scenario}} & \multicolumn{4}{c}{\textbf{Average Performance}} \\
        \cline{2-5}
                                           & \textbf{k=1} & \textbf{k=3} & \textbf{k=5} & \textbf{k=10} \\
        \hline
        Greeting                           & 1.4706 & 1.5229 & \textbf{2.0867} & 1.6209 \\
        \hline
        Dialogue                           & 1.7829 & 1.8022 & \textbf{2.2058} & 1.8646 \\
        \bottomrule
    \end{tabular}
\end{table}

\subsection{Online Experimental Result}

To evaluate the system’s impact on user engagement, we conduct an online A/B test. We select Average Dialogue Duration as the primary evaluation metric, as it directly reflects user engagement by measuring the average time users spend in dialogue interactions. To establish a baseline, we employ chatbots with passive dialogue strategies. The experiment is conducted over a 7-day period, with participants evenly assigned to the control and experimental groups in a 1:1 ratio. As shown in Table \ref{tab:ab}, compared to the baseline, PaRT achieves a 21.77\% increase in average dialogue duration. This result indicates that our approach effectively enhances user engagement and facilitates more in-depth conversations.

\begin{table}[htbp]
    \caption{A/B Test Result}
    \label{tab:ab}
    \centering
    \begin{tabular}{p{3.8cm}cc}
        \toprule
        \textbf{Metric} & \textbf{Baseline} & \textbf{PaRT}  \\
        \hline
        Average Dialogue Duration (s) & 296.88 & \textbf{361.51 (+21.77\%)}\\
        \bottomrule
    \end{tabular}
\end{table}

\section{Conclusion}
In this paper, we propose PaRT, a method that empowers social chatbots with proactive dialogue and real-time information enrichment. 
By integrating user profiling, intent-guided query refiner, and retrieval-augmented generation, it enables human-like conversation initiation and smooth topic transitions.
Experimental results confirm improved dialogue quality and increased user engagement.

\newpage

\section*{Speaker Biography}
\textbf{Zihan Niu} is a master's student at the University of Science and Technology of China. His research focuses on retrieval-augmented generation and conversational AI agents. He has published papers at SIGIR, EMNLP, amongst others.

\bibliographystyle{ACM-Reference-Format}
\bibliography{reference}

\end{document}